\documentclass[letterpaper, 10 pt, journal, twoside]{IEEEtran}

\IEEEoverridecommandlockouts                              

\usepackage{amsmath}
\usepackage[usenames,dvipsnames]{xcolor}
\usepackage{graphicx} 
\usepackage[acronym]{glossaries}
\usepackage{hyperref}
\usepackage[caption=false,font=footnotesize]{subfig}
\usepackage{todonotes}
\usepackage{cite} 
\usepackage{float}
\usepackage{multirow}
\usepackage{booktabs}
\usepackage{array}
\usepackage{enumitem} 

\usepackage{xcolor,colortbl}
\usepackage{color,soul}
\usepackage{rotating}
\definecolor{bedColor}{rgb}{0, 0, 1}
\definecolor{chairColor}{RGB}{192, 64, 64}
\definecolor{sofaColor}{rgb}{0, 0.8549, 0}
\definecolor{tableColor}{RGB}{192,128,192}
\definecolor{booksColor}{rgb}{0.8706,0.9451,0.0941}
\definecolor{refrigeratorColor}{RGB}{64, 192, 0}
\definecolor{tvColor}{RGB}{0, 64, 192}
\definecolor{bagColor}{RGB}{128, 64, 192}
\definecolor{toiletColor}{rgb}{0.4588,0.1137,0.1608}

\usepackage{dingbat}
\usepackage{tabularx}
\usetikzlibrary{backgrounds}
\usetikzlibrary{positioning}

\usepackage{textcomp}

\newcolumntype{C}[1]{>{\centering\arraybackslash}p{#1}}

\title{Volumetric Instance-Aware Semantic Mapping and 3D Object Discovery}

\author{Margarita Grinvald, Fadri Furrer, Tonci Novkovic, Jen Jen Chung, Cesar Cadena, Roland Siegwart, Juan Nieto%
\thanks{This work was supported by ABB Corporate Research, the Amazon Research Awards program, and the Swiss National Science Foundation (SNF) through the National Centre of Competence in Research on Digital Fabrication. \textit{(Corresponding author: Margarita Grinvald.)}}
\thanks{The authors are with the Autonomous Systems Lab, ETH Zurich, 8092 Zurich, Switzerland 
 (e-mail: mgrinvald@ethz.ch; fadri@ethz.ch; ntonci@ethz.ch; chungj@ethz.ch; cesarc@ethz.ch; rsiegwart@ethz.ch; nietoj@ethz.ch).}
\thanks{Digital Object Identifier 10.1109/LRA.2019.2923960}}

\newcommand{\acronym}[1]{\gls{#1}\@}

\newacronym{cnn}{CNN}{Convolutional Neural Network}
\newacronym{tsdf}{TSDF}{Truncated Signed Distance Field}
\newacronym{cnns}{CNNs}{Convolutional Neural Network}
\newacronym{sdf}{SDF}{Signed Distance Field}
\newacronym{slam}{SLAM}{Simultaneous Localization and Mapping}
\newacronym{dof}{DoF}{Degrees of Freedom}
\newacronym{iodb}{IODB}{Incremental Object Database}
\newacronym{ap}{AP}{Average Precision}
\newacronym{map}{mAP}{mean Average Precision}
\newacronym{iou}{IoU}{Intersection over Union}

\begin{document}


\maketitle

\IEEEpubid{\begin{minipage}{\textwidth}\ \\[12pt] \centering
  2377-3766 \copyright~2019 IEEE. Personal use is permitted, but republication/redistribution requires IEEE permission. \\
  See http://www.ieee.org/publications\_standards/publications/rights/index.html for more information.
\end{minipage}} 

\begin{abstract}
To autonomously navigate and plan interactions in real-world environments, robots require the ability to robustly perceive and map complex, unstructured surrounding scenes. 
Besides building an internal representation of the observed scene geometry, the key insight toward a truly functional understanding of the environment is the usage of higher-level entities during mapping, such as individual object instances.
This work presents an approach to incrementally build volumetric object-centric maps during online scanning with a localized \mbox{RGB-D} camera.
First, a per-frame segmentation scheme combines an unsupervised geometric approach with instance-aware semantic predictions to detect both recognized scene elements as well as previously unseen objects.
Next, a data association step tracks the predicted instances across the different frames.
Finally, a map integration strategy fuses information about their 3D shape, location, and, if available, semantic class into a global volume.
%
Evaluation on a publicly available dataset shows that the proposed approach for building instance-level semantic maps is competitive with state-of-the-art methods, while additionally able to discover objects of unseen categories.
The system is further evaluated within a real-world robotic mapping setup, for which qualitative results highlight the online nature of the method.
Code is available at \url{https://github.com/ethz-asl/voxblox-plusplus}.
\end{abstract}


\section{INTRODUCTION}
\IEEEPARstart{R}{obots} operating autonomously in unstructured, real-world environments cannot rely on a detailed \textit{a priori} map of their surroundings for planning interactions with scene elements.
They must therefore be able to robustly perceive the complex surrounding space and acquire task-relevant knowledge to guide subsequent actions. 
Specifically, to learn accurate 3D object models for tasks such as grasping and manipulation, a robotic vision system should be able to discover, segment, track, and reconstruct objects at the level of the individual instances.
\IEEEpubidadjcol
However, real-world scenarios exhibit large variability in object appearance, shape, placement, and location, posing a direct challenge to robotic perception.
Further, such settings are usually characterized by open-set conditions, i.e.\ the robot will inevitably encounter novel objects of previously unseen categories.

Computer vision algorithms have shown impressive results for the tasks of detecting individual objects in RGB images and predicting for each a per-pixel semantically annotated mask~\cite{MaskRCNN,2018SceneCutBayesian}.
%
On the other hand, dense 3D scene reconstruction has been extensively studied by the robotics community.
Combining the two areas of research, a number of works successfully locate and segment semantically meaningful objects in reconstructed scenes while dealing with substantial intra-class variability~\cite{sunderhauf17,maskfusion,PhamInstances,FusionPP}.
%
%
%
%
%
\IEEEpubidadjcol
\begin{figure}[!t]
\renewcommand{\arraystretch}{0.9}
\centering
\setlength{\tabcolsep}{0.2\tabcolsep}
\begin{tabular}{cc}
\subfloat[Object-centric Map\label{fig:gim1}]{\begin{tikzpicture}[every node/.style={inner sep=0,outer sep=0}, tight background]
\node(a){\includegraphics[width=0.48\columnwidth]{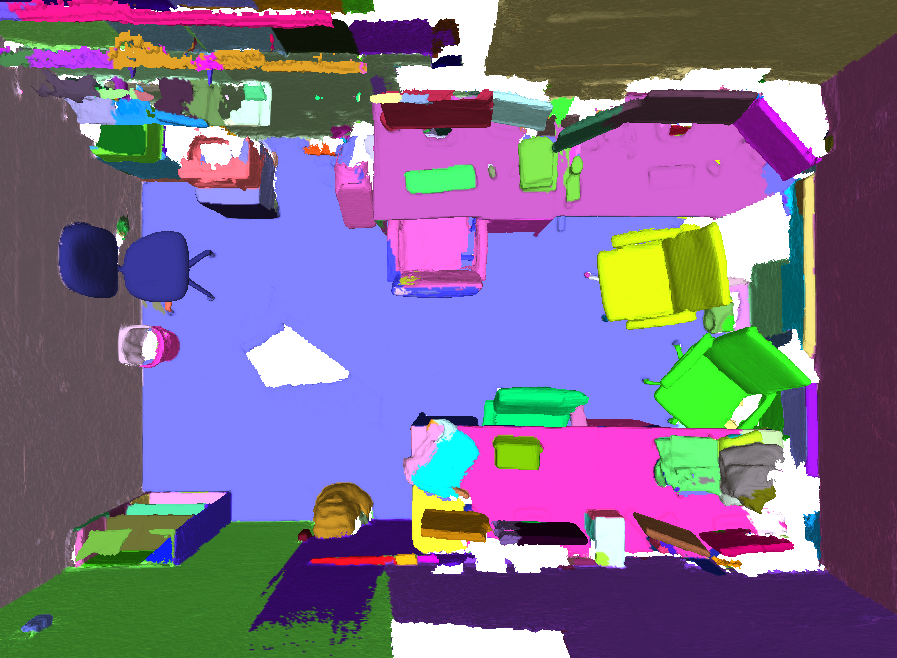}};
\node at(a.center) [
    circle,
    draw,
    red,
    very thick,
    minimum size=5ex,
    yshift=-16.5pt,
    xshift=3pt
] {};
\node at(a.center) [
    circle,
    draw,
    blue,
    very thick,
    minimum size=5ex,
    yshift=6.9pt,
    xshift=28pt
] {};
\end{tikzpicture}} &
\subfloat[Ground Truth Instance Map\label{fig:gt1}]{\begin{tikzpicture}[every node/.style={inner sep=0,outer sep=0}, tight background]
\node(b){\includegraphics[width=0.48\columnwidth]{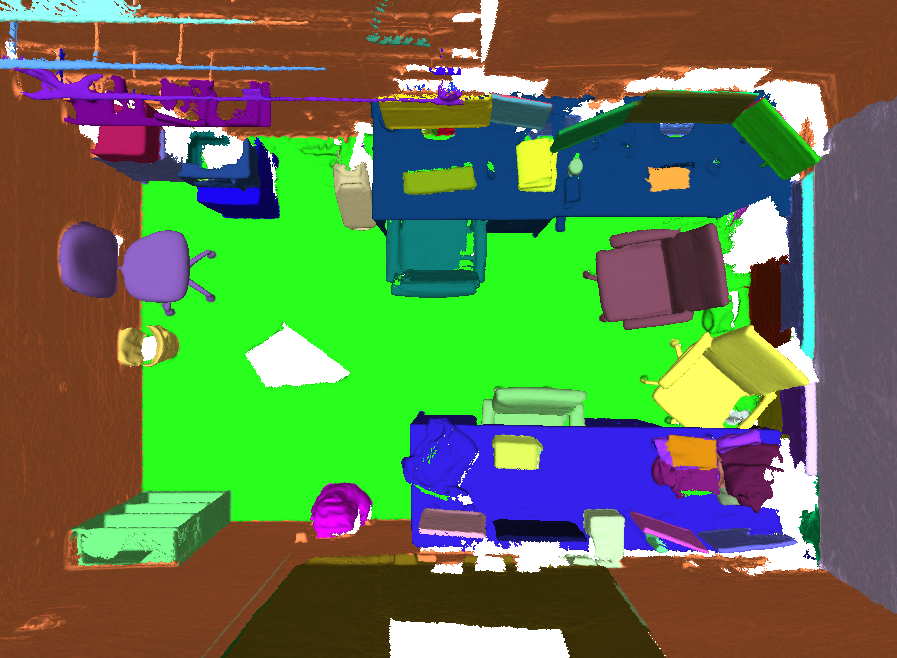}};
\end{tikzpicture}} \\[-2mm]
\subfloat[Semantic Instance Segmentation\label{fig:semantic1}]{\begin{tikzpicture}[every node/.style={inner sep=0,outer sep=0}, tight background]
\node(c){\includegraphics[width=0.48\columnwidth]{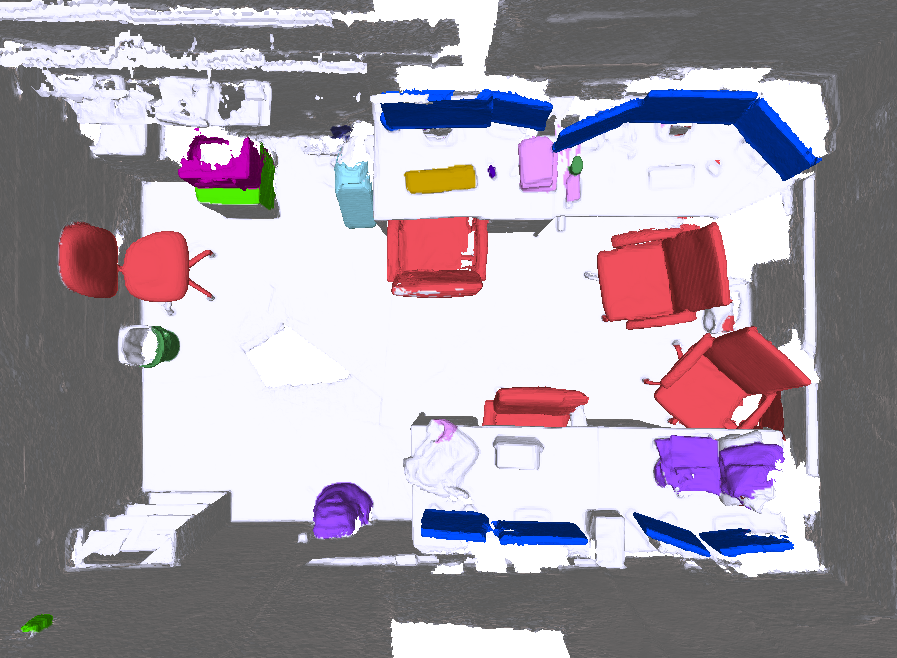}};
\node at(c.center) [
    circle,
    draw,
    red,
    very thick,
    minimum size=5ex,
    yshift=-16.5pt,
    xshift=3pt
] {};
\end{tikzpicture}} &
\subfloat[Geometric Segmentation~\cite{Furrer2018IncrementalOD}\label{fig:gsm1}]{\begin{tikzpicture}[every node/.style={inner sep=0,outer sep=0}, tight background]
\node(d){\includegraphics[width=0.48\columnwidth]{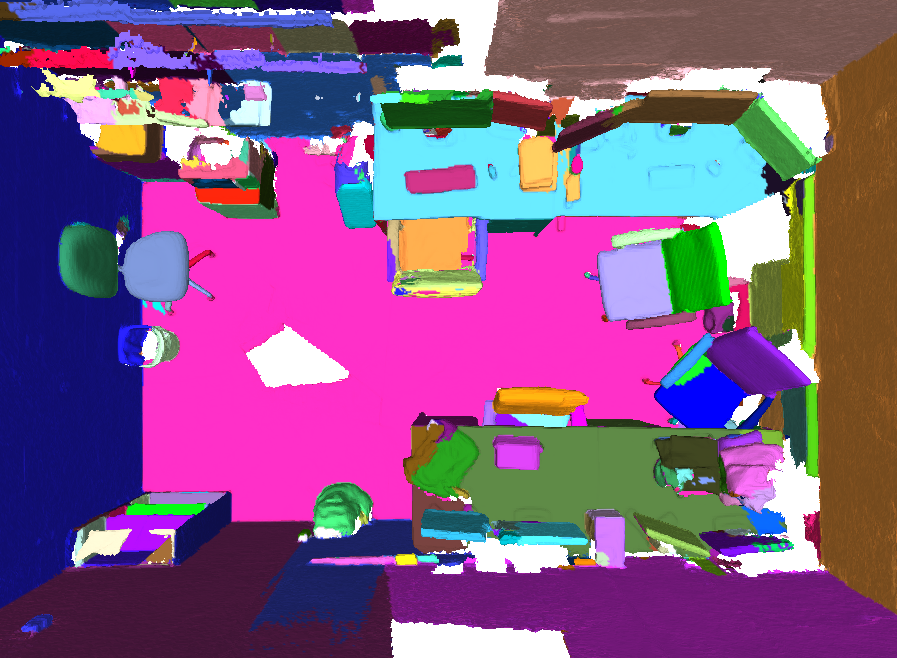}};
\node at(d.center) [
    circle,
    draw,
    blue,
    very thick,
    minimum size=5ex,
    yshift=6.9pt,
    xshift=28pt
] {};
\end{tikzpicture}} \\\\[-2mm]
\multicolumn{2}{c}{
\centering
\begingroup
\footnotesize
\begin{tabularx}{\columnwidth}{XXXX}
\tikz\node[rectangle, minimum width=12pt, fill={rgb,255:red,0;green,64;blue,192}]{}; Monitor & 
\tikz\node[rectangle, minimum width=11pt, fill={rgb,255:red,160;green,128;blue,0}]{}; Keyboard  & 
\tikz\node[rectangle, minimum width=11pt, fill={rgb,255:red,128;green,192;blue,192}]{}; Suitcase &
\tikz\node[rectangle, minimum width=11pt, fill={rgb,255:red,192;green,128;blue,192}]{}; Table \\
\tikz\node[rectangle, minimum width=11pt, fill={rgb,255:red,192;green,64;blue,64}]{}; Chair & 
\tikz\node[rectangle, minimum width=11pt, fill={rgb,255:red,64;green,0;blue,128}]{}; Mouse &
\tikz\node[rectangle, minimum width=11pt, fill={rgb,255:red,64;green,192;blue,0}]{}; Refrigerator & 
\tikz\node[rectangle, minimum width=11pt, fill={rgb,255:red,64;green,64;blue,128}]{}; Plant\\
\tikz\node[rectangle, minimum width=11pt, fill={rgb,255:red,128;green,64;blue,192}]{}; Backpack &
\tikz\node[rectangle, minimum width=11pt, fill={rgb,255:red,64;green,128;blue,64}]{}; Cup  &
\tikz\node[rectangle, minimum width=11pt, fill={rgb,255:red,160;green,0;blue,128}]{}; Microwave &
\tikz\node[rectangle, minimum width=11pt, fill={rgb,255:red,200;green,200;blue,200}]{}; Unknown 
\end{tabularx}%
\endgroup
}
\vspace{-2mm}
\end{tabular}
\caption{Reconstruction and object-level segmentation of an office scene using the proposed approach.
Besides accurately describing the observed surface geometry, the final object-centric map in Figure~\protect\subref{fig:gim1} carries information about the location and 3D shape of the individual object instances in the scene. 
As opposed to a geometry-only segmentation from our previous work\mbox{\protect\cite{Furrer2018IncrementalOD}} shown in Figure~\mbox{\protect\subref{fig:gsm1}}, the proposed framework prevents over-segmentation of recognized articulated objects and segments them as one instance despite their non-convex shape (blue circle), assigning each a semantic category shown in Figure~\mbox{\protect\subref{fig:semantic1}}.
At the same time, the proposed approach discovers novel, previously unseen object-like elements of unknown class (red circle).
Note that different colors in Figure~\protect\subref{fig:gim1} and Figure~\protect\subref{fig:gt1} represent the different instances, and that a same instance in the prediction and ground truth is not necessarily of the same color.
Progressive mapping of sequence 231 from the SceneNN~\protect\cite{7785081} dataset is shown in the accompanying video available at \protect\url{http://youtu.be/Jvl42VJmYxg}.}
\label{fig:teaser}
\vspace{-6mm}
\end{figure}
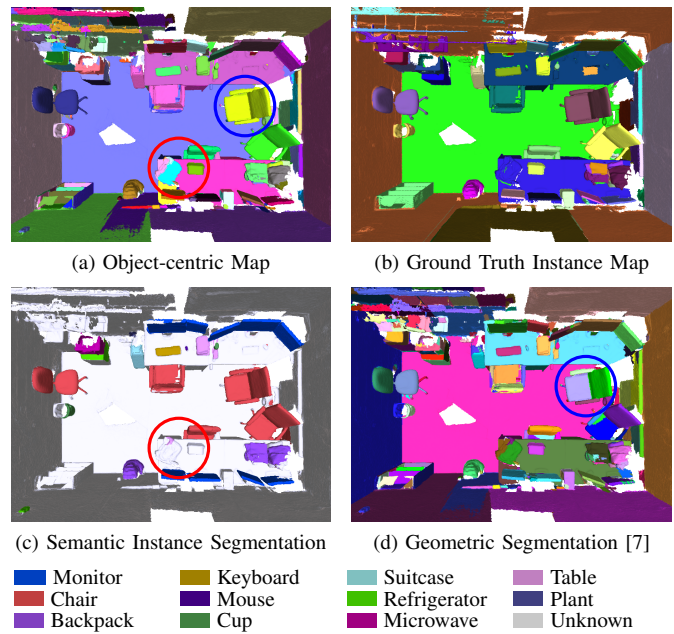
Still, these methods can only detect objects from a fixed set of classes used during training, thus limiting interaction planning to a subset of the observed elements.
In contrast, purely geometry-based methods~\cite{Furrer2018IncrementalOD,7354011} are able to discover novel, previously unseen scene elements, under open-set conditions.
However, such approaches tend to over-segment the reconstructed objects and additionally fail to provide any semantic information about them, making high-level scene understanding and task planning impractical.

This letter presents an approach to incrementally build geometrically accurate volumetric maps of the environment that additionally contain information about the individual object instances observed in the scene.
In particular, the proposed object-oriented mapping framework retrieves the pose and shape of recognized semantic objects, as well as of newly discovered, previously unobserved object-like instances.
The proposed system builds on top of the incremental \mbox{geometry-based} scene segmentation approach from our previous work in~\cite{Furrer2018IncrementalOD} and extends it to produce a complete instance-aware semantic mapping framework.
Figure~\ref{fig:teaser} shows the object-centric map of an office scene reconstructed with the proposed approach.

The system takes as input the RGB-D stream of a depth camera with known pose.\footnote{
Please note that the current work focuses entirely on mapping, hence localization of the camera is assumed to be given.}
First, a frame-wise segmentation scheme combines an unsupervised geometric segmentation of depth images~\cite{7354011} with semantic object predictions from RGB~\cite{MaskRCNN}.
The use of semantics allows the system to infer the category of some of the 3D segments predicted in a frame, as well as to group segments by the object instance to which they belong.
Next, the tracking of the individual predicted instances across multiple frames is addressed by matching per-frame predictions to existing segments in the global map via a data association strategy.
Finally, observed surface geometry and segmentation information are integrated into a global \acronym{tsdf} map volume.
To this end, the Voxblox volumetric mapping framework~\cite{oleynikova2017voxblox} is extended to enable the incremental fusion of class and instance information within the reconstruction.
By relying on a volumetric representation that explicitly models free space information, i.e.\ distinguishes between unknown space and observed, empty space, the built maps can be directly used for safe robotic navigation and motion planning purposes.
Furthermore, object models reconstructed with the voxel grid explicitly encode surface connectivity information, relevant in the context of robotic manipulation applications.

The capabilities of the proposed method are demonstrated in two experimental settings.
First, the proposed instance-aware semantic mapping framework is evaluated on office sequences from the real-world SceneNN~\cite{7785081} dataset to compare against previous work on progressive instance segmentation of 3D scenes.
Lastly, we show qualitative results for an online mapping scenario on a robotic platform.
The experiments highlight the robustness of the presented incremental segmentation strategy, and the online nature of the framework.

The main contributions of this work are:
\begin{itemize}[noitemsep,topsep=0pt]
    \item A combined geometric-semantic segmentation scheme that extends object detection to novel, previously unseen categories.
    \item A data association strategy for tracking and matching instance predictions across multiple frames.
    \item Evaluation of the framework on a publicly available dataset and within an online robotic mapping setup.
\end{itemize}

\section{RELATED WORK}

\subsection{Object detection and segmentation}
In the context of object recognition in real-world environments, computer vision algorithms have recently shown some impressive results.
Driven by the advances in deep learning using \acronym{cnns}, several architectures have been proposed for detecting objects in RGB images~\cite{Redmon2017YOLO9000BF, FasterRCNN}. 
Beyond simple bounding boxes, the recent Mask R-CNN framework~\cite{MaskRCNN} is further able to predict a per-pixel semantically annotated mask for each of the detected instances, achieving state-of-the-art results on the COCO instance-level semantic segmentation task~\cite{COCO}.

%
%

One of the major limitations of learning-based instance segmentation methods is that they require extensive amounts of training data in the form of annotated masks for the specified object categories.
Such annotated data can be expensive or even infeasible to acquire for all possible categories that may be encountered in a real-world scenario.
Moreover, these algorithms can only recognize the fixed set of classes provided during training, thus failing to correctly segment and classify other, previously unseen object categories.

Some recent works aim to relax the requirement for large amounts of pixel-wise semantically annotated training data.
Mask$^X$~R-CNN~\cite{MaskXRCNN} adopts a transfer method which only requires a subset of the data to be labeled at training time.
SceneCut~\cite{SceneCut} and its Bayesian extension in~\cite{2018SceneCutBayesian} also operate under open-set conditions and are able to detect and segment novel objects of unknown classes.
However, beyond detecting object instances in individual image frames, these methods alone do not provide a comprehensive 3D representation of the scene and, therefore, cannot be directly used for planning tasks such as manipulation or navigation.

\subsection{Semantic object-level mapping}
Recent developments in deep learning have also enabled the integration of rich semantic information within real-time \acronym{slam} systems.
The work in~\cite{SemanticFusion} fuses semantic predictions from a CNN into a dense map built with a \acronym{slam} framework.
However, conventional semantic segmentation is unaware of object instances, i.e.\ it does not disambiguate between individual instances that belong to the same category.
Thus, the approach in~\cite{SemanticFusion} does not provide any information about the geometry and relative placement of individual objects in the scene.
Similar work in \cite{FastAccurateSemanticMapping} additionally proposes to incrementally segment the scene using geometric cues from depth.
However, geometry-based approaches tend over-segment articulated scene elements.
Thus, without instance-level information, a joint semantic-geometric segmentation is not enough to group parts of the scene into distinct separate objects. 
Indeed, the instance-agnostic semantic segmentation in these works fails to build semantically meaningful maps to model individual object instances.

Previous work has addressed the task of mapping at the level of individual objects.
SLAM++~\cite{6619022} builds object-oriented maps by detecting recognized elements in RGB-D data, but is limited to work with a database of objects for which exact geometric models need to be known in advance.
A number of other works have addressed the task of detecting and segmenting individual semantically meaningful objects in 3D scenes without predefined shape templates~\cite{Furrer2018IncrementalOD,7354011,sunderhauf17,maskfusion,PhamInstances, FusionPP}. 
Recent learning-based approaches segment individual instances of semantically annotated objects in reconstructed scenes with little or no prior information about their exact appearance while at the same time handling substantial intra-class variability \cite{sunderhauf17,maskfusion,PhamInstances, FusionPP}.
However, by relying on a strong supervisory signal of the predefined classes during training, a purely learning-based segmentation fails to discover novel objects of unknown class in the scene.
As a result, these methods either fail to map objects that do not belong to the set of known categories and for which no semantic labels are predicted~\cite{sunderhauf17,maskfusion, FusionPP}, or wrongly assign such previously unseen instances to one of the known classes~\cite{PhamInstances}.
In a real-world scenario, detecting objects only from a fixed set of classes specified during training limits interaction planning to a subset of all the observed scene elements.

\begin{figure*}[ht]
    \centering
    \includegraphics[width=0.85\textwidth]{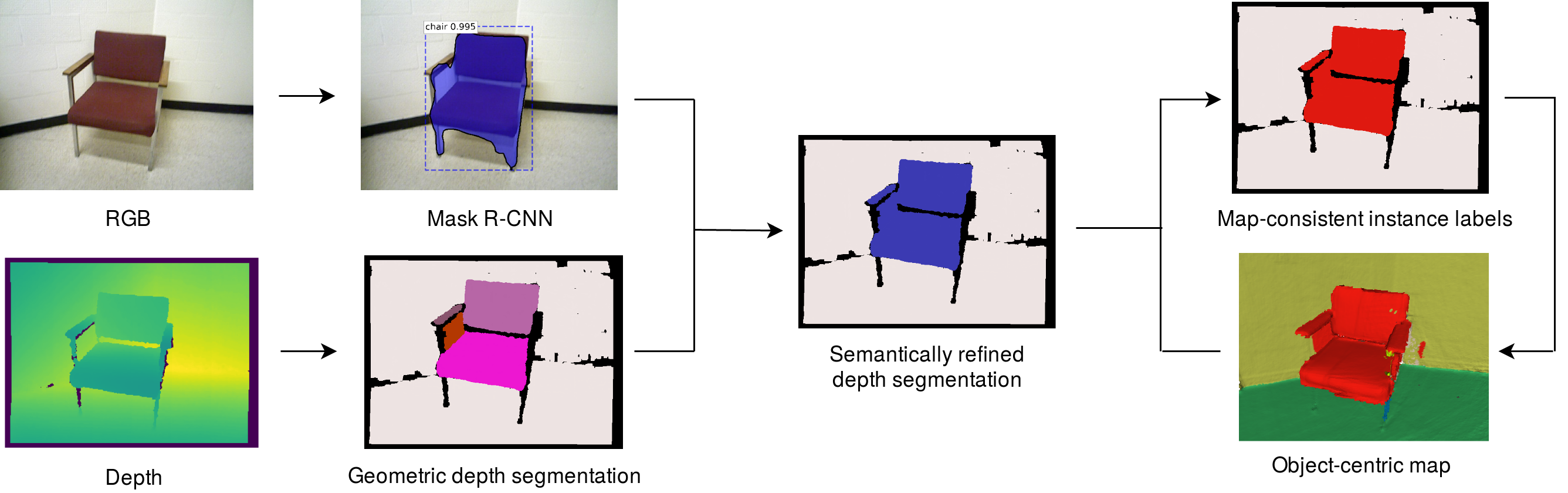}
    \caption{The individual stages of the proposed approach for incremental object-level mapping are illustrated here with an example.
At each new frame, the incoming RGB image is processed with the Mask R-CNN network to detect object instances and predict for each a semantically annotated mask.
At the same time, a geometric segmentation decomposes the depth image into a set of convex 3D segments.
The predicted semantic mask is used to infer class information for the corresponding depth segments and to refine over-segmentation of non-convex objects by grouping segments by the object instance they belong to.
Next, a data association strategy matches segments predicted in the current frame to their corresponding instance in the global map to retrieve for each a map-consistent label. 
Finally, dense geometry and segmentation information from the current frame are integrated into the global map volume.}
\label{fig:overview}
\vspace{-4mm}
\end{figure*}

In contrast, purely geometry-based methods operate under open-set conditions and are able to discover novel, previously unobserved objects in the scene~\cite{7354011,Furrer2018IncrementalOD}.
The work in~\cite{7354011} provides a complete and exhaustive geometric segmentation of the scene.
Similarly, the \acronym{iodb} in~\cite{Furrer2018IncrementalOD} performs a purely geometric segmentation from depth data to reconstruct the shape of individual segments and build a consistent database of unique 3D object models.
However, as mentioned previously, geometry-based approaches can result in unwanted over-segmentation of non-convex objects.
Furthermore, by not providing semantic information, the two methods disallow high-level interaction planning.
In addition to a complete geometric segmentation of the scene, the work in~\cite{7487378} performs object recognition on such segments from a database of known objects.
While able to discover new, previously unseen objects and to provide for some semantic information, the main drawback lies in the requirement for exact 3D geometric models of the recognized objects to be known.
This is not applicable to real-world environments, where objects with novel shape variations are inevitably encountered on a regular basis.

Closely related to the approach presented in this letter is the recent work in~\cite{NakajimaInstance}, with the similar aim of building dense object-oriented semantic maps.
The work presents an incremental geometry-based segmentation strategy, coupled with the YOLO~v2~\cite{Redmon2017YOLO9000BF} bounding box detector to identify and merge geometric segments that are detected as part of the same instance.
One of the key differences to our approach is the choice of scene representation.
Their system relies on the RGB-D \acronym{slam} system from~\cite{InfiniTAM_V3_Report_2017} and stores the reconstructed 3D map through a surfel-based representation~\cite{KellerSurfels}.
While surfels allow for efficient handling of loop closures, they only store the surface of the environment and do not explicitly represent observed free space~\cite{WurmOctrees}.
That is, a surfel map does not distinguish between unseen and seen-but-empty space, and thus cannot be directly used for planning in robotic navigation or manipulation tasks where knowledge about free space is essential for safe operation~\cite{VespaOctree}.
Further, visibility determination and collision detection in surfel clouds can be significantly harder due to the lack of surface connectivity information.
Therefore, as with all other approaches relying on sparse point or surfel clouds representations~\cite{sunderhauf17, maskfusion}, the object-oriented maps built in~\cite{NakajimaInstance} cannot be immediately used in those robotic settings where an explicit distinction between unobserved space and free space is required.

Conversely, the volumetric \acronym{tsdf}-based representation adopted in this work does not discard valuable free space information and explicitly distinguishes observed empty space from unknown space in the 3D map.
In contrast to all previous approaches, the proposed method is able to incrementally provide densely reconstructed volumetric maps of the environment that contain shape and pose information about both recognized and unknown object elements in the scene.
The reconstructed maps are expected to directly benefit navigation and interaction planning applications.

\section{METHOD}
The proposed incremental object-level mapping approach consists of four steps deployed at each incoming RGB-D frame: (i)~geometric segmentation, (ii)~semantic instance-aware segmentation refinement, (iii)~data association, and (iv)~map integration.
First, the incoming depth map is segmented according to a convexity-based geometric approach that yields segment contours which accurately describe real-world physical boundaries (Section~\ref{ssec:depth_segmentation}).
The corresponding RGB frame is processed with the Mask R-CNN framework to detect object instances and compute for each a per-pixel semantically annotated segmentation mask.
The per-instance masks are used to semantically label the corresponding depth segments and to merge segments detected as belonging to the same geometrically over-segmented, non-convex object instance (Section~\ref{ssec:seg_refinement}).
A data association strategy matches segments discovered in the current frame and their comprising instances to the ones already stored in the map (Section~\ref{ssec:data_association}).
Finally, segments are integrated into the dense 3D map, where a fusion strategy keeps track of the individual segments discovered in the scene (Section~\ref{ssec:integration}).
An example illustrating the individual stages of the proposed approach is shown in Figure~\ref{fig:overview}.

\subsection{Geometric segmentation}
\label{ssec:depth_segmentation}
Building on the assumption that real-world objects exhibit overall convex surface geometries, each incoming depth frame is decomposed into a set of object-like convex 3D segments following the geometry-based approach introduced in~\cite{Furrer2018IncrementalOD}.
First, surface normals are estimated at every depth image point. 
Next, angles between adjacent normals are compared to identify concave region boundaries.
Additionally, large 3D distances between adjacent depth map vertices are used to detect strong depth discontinuities.
Surface convexity and the 3D distance measure are then combined to generate, at every frame $t$, a set $\mathcal{R}_t$ of closed 2D regions $r_i$ in the current depth image and a set $\mathcal{S}_t$ of corresponding 3D segments $s_i$.
Figure~\ref{fig:overview} shows the sample output of this stage.

\subsection{Semantic instance-aware segmentation refinement}
\label{ssec:seg_refinement}
To complement the unsupervised geometric segmentation of each depth frame with semantic object instance information, the corresponding RGB images are processed with the Mask~R-CNN framework~\cite{MaskRCNN}.
The network detects and classifies individual object instances and predicts a semantically annotated segmentation mask for each of them.
Specifically, for each input RGB frame the output is a set of object instances, where the $k$-th detected instance is characterized by a binary mask $M_k$ and an object category $c_k$.
Figure~\ref{fig:overview} shows the sample output of Mask R-CNN.

The segmentation masks offer a straightforward way to associate each of the detected instances with one or more corresponding 3D depth segments $s_i\in \mathcal{S}_t$.
Pairwise 2D overlaps ${p}_{i, k}$ between each $r_i\in \mathcal{R}_t$ and each predicted binary mask $M_k$ are computed as the number of pixels in the intersection of $r_i$ and $M_k$ normalized by the area of $r_i$:

\begin{equation}
p_{i, k} = \cfrac{|r_i \cap M_k|}{|r_i|}\enspace .
\end{equation}
For each region $r_i\in\mathcal{R}_t$ the highest overlap percentage $p_i$ and index $\hat{k}_i$ of the corresponding mask $M_k$ are found as:
\begin{gather}
p_{i} = \max_{k} ~ p_{i, k} \enspace \\
\hat{k}_i = \arg\max_{k} ~ p_{i, k} \enspace.
\end{gather}
If $p_{i} >\tau_p$, the corresponding 3D segment $s_i$ is assigned the object instance label $o_i = \hat{k}_i$ and a semantic category $c_i = c_{\hat{k}_i}$.
%
Multiple segments in $\mathcal{S}_t$ assigned to the same object instance label $o_i$ value indicate an over-segmentation of non-convex, articulated shapes being refined through semantic instance information.
The unique set of all object instance labels $o_i$ assigned to segments $s_i\in \mathcal{S}_t$ in the current frame is denoted by $\mathcal{O}_t$.
All segments $s_i\in\mathcal{S}_t$ for which no mask $M_k$ in the current frame exhibits enough overlap are assigned $o_i = c_i = 0$, denoting a geometric segment for which no semantic instance information could be predicted.

\renewcommand{\arraystretch}{1.2}
\begin{table*}[ht]
\caption{Comparison to the 3D semantic instance-segmentation approach from Pham \textit{et al.}~\cite{PhamInstances}. Per-class AP is evaluated using an IoU threshold of 0.5 for each of the 10 evaluated sequences from the SceneNN~\cite{7785081} dataset. The class-averaged mAP value is compared to the results presented in~\cite{PhamInstances}. The proposed approach improves over the baseline for 7 of the 10 sequences evaluated, however it is worth noting that the reported mAP values are evaluated on a smaller set of classes compared to the ones from~\cite{PhamInstances}.}
\centering
\begin{tabular}{c|C{0.5cm}C{0.5cm}C{0.5cm}C{0.5cm}C{0.5cm}C{0.5cm}C{0.5cm}C{0.5cm}C{0.5cm}|C{0.5cm}|C{0.5cm}}
Sequence ID  & 
\multicolumn{1}{c|}{\cellcolor{bedColor}\begin{turn}{90}\color{white}Bed\end{turn}} & 
\multicolumn{1}{c|}{\cellcolor{chairColor}\begin{turn}{90}\color{white}Chair\end{turn}} & 
\multicolumn{1}{c|}{\cellcolor{sofaColor}\begin{turn}{90}Sofa\end{turn}} & \multicolumn{1}{c|}{\cellcolor{tableColor}\begin{turn}{90} Table\end{turn}} & \multicolumn{1}{c|}{\cellcolor{booksColor}\begin{turn}{90} Books\end{turn}} & \multicolumn{1}{c|}{\cellcolor{refrigeratorColor}\begin{turn}{90} Refrigerator\end{turn}} & \multicolumn{1}{c|}{\cellcolor{tvColor}\begin{turn}{90} \color{white}Television\end{turn}} & \multicolumn{1}{c|}{\cellcolor{toiletColor}\begin{turn}{90}\color{white}Toilet\end{turn}}  & \multicolumn{1}{c|}{\cellcolor{bagColor}\begin{turn}{90} \color{white}Bag\end{turn}} & \multicolumn{1}{c|}{\begin{turn}{90} Average\end{turn}} & \multicolumn{1}{c}{\begin{turn}{90} Pham \textit{et al.}~\cite{PhamInstances}\end{turn}} \\ 
\midrule
\multicolumn{1}{c|}{011} & - & 75.0 & 50.0 & 100 & - & - & -  & -  & - & \textbf{75.0} & 52.1 \\
\multicolumn{1}{c|}{016} & 100 & 0.0 & 0.0  & - & - & - & -  & -  & - &  33.3 & \textbf{34.2}\\
\multicolumn{1}{c|}{030} & - & 54.4 & 100  & 55.6 & 14.3 & - & -  & -  & - & 56.1 &  \textbf{56.8}\\
\multicolumn{1}{c|}{061} & - & - & 100  & 33.3 & - & - & -  & -  & - &  \textbf{66.7} & 59.1\\
\multicolumn{1}{c|}{078} & - & 33.3 & -  &  0.0  & 47.6 & 100 & -  & -  & - & \textbf{45.2} &  34.9\\
\multicolumn{1}{c|}{086} & - & 80.0 & -  & 0.0 & 0.0 & - & -  & -  & 0.0 & 20.0 &  \textbf{35.0}\\
\multicolumn{1}{c|}{096} & 0.0 & 87.5 & -  & 37.5 & 0.0 & - & 0.0  & -  & 50 & \textbf{29.2} &  26.5 \\
\multicolumn{1}{c|}{206} & - & 58.3 & 100  & 60.0 & - & - & -  & -  & 100 & \textbf{79.6} &  41.7 \\
\multicolumn{1}{c|}{223} & - & 12.5 & -  & 75.0 & - & - & -  & -  & - & \textbf{43.8} &  40.9\\
\multicolumn{1}{c|}{255} & - & - & -  & - & - & 75.0 & -  & -  & - & \textbf{75.0} &   48.6 \\
\end{tabular}
\label{tab:evaluation}
\vspace{-4mm}
\end{table*}

\subsection{Data association}
\label{ssec:data_association}
Because the frame-wise segmentation processes each incoming RGB-D image pair independently, it lacks any spatio-temporal information about corresponding segments and instances across the different frames.
Specifically, this means that it does not provide an association between the set of predicted segments $\mathcal{S}_t$ and the set of segments $\mathcal{S}_{t+1}$.
Further, segments belonging to the same object instance might be assigned different $o_i$ label values across two consecutive frames, since these represent mask indices valid only within the scope of the frame in which such masks were predicted.

A data association step is proposed here to track corresponding geometric segments and predicted object instances across frames.
To this end, we define a set of persistent geometric labels $\mathcal{L}$ and a set of persistent object instance labels $\mathcal{O}$ which remain valid throughout the entire mapping session.
In particular, each $s_j$ from the set of segments $\mathcal{S}$ stored in the map is defined by a unique geometric label $l_j\in\mathcal{L}$ through a mapping $L(s_j) = l_j$.
At each frame we then look for a mapping $L_t(s_i) = l_j$ that matches predicted segments $s_i\in\mathcal{S}_t$ to corresponding segments $s_j\in\mathcal{S}$.
Similarly, within the scope of a frame we seek to define a mapping $I_t(o_i) = o_m$ that matches object instances $o_i\in\mathcal{O}_t$ to persistent instance labels $o_m\in\mathcal{O}$ stored in the map.

To track spatial correspondences between segments $s_i\in\mathcal{S}_t$ identified in the current depth map and the set $\mathcal{S}$ of segments in the global map it is only necessary to consider the set $\mathcal{S}_v \subset \mathcal{S}$ of map segments visible in the current camera view.
%
The pairwise 3D overlap $\Pi_{i,j}$ is computed for each $s_i\in\mathcal{S}_t$ and each $s_j\in\mathcal{S}_v$ as the number of points in segment $s_i$ that, when projected into the global map frame using the known camera pose, correspond to a voxel which belongs to segment $s_j$.
For each segment $s_j\in\mathcal{S}_v$, the highest overlap measure $\Pi_{j}$ and the index $\hat{i}_j$ of the corresponding segment $s_i\in\mathcal{S}_t$ are found as,
\begin{gather}
\Pi_{j} = \max_{i} ~ \Pi_{i,j} \enspace \\
\hat{i}_j = \arg\max_{i} ~ \Pi_{i,j} \enspace.
\end{gather}
Each segment $s_j\in\mathcal{S}_v$ with $\Pi_{j}>\tau_{\pi}$ determines the persistent label mapping for the corresponding maximally overlapping segment $s_{\hat{i}_j}\in\mathcal{S}_t$ from the current depth frame, i.e.\ $L_t(s_{\hat{i}_j}) = L(s_j)$.
The $\tau_{\pi}$ threshold value is set to 20, and is used to prevent poorly overlapping global map segment labels from being propagated to the current frame.
All segments $s_i\in\mathcal{S}_t$ that did not match to any segment $s_j\in\mathcal{S}_v$ are assigned a new persistent label $l_{new}$ as $L_t(s_i) = l_{new}$.
It is worth noting that, in contrast to previous work on segment tracking across frames\mbox{\protect\cite{7354011}}, the proposed formulation disallows matching multiple segments in  $\mathcal{S}_t$ to the same segment $s_j\in\mathcal{S}_v$.
%
Without such constraint, information about a region in the map that was initially segmented as one now being segmented in two or more parts in the current frame would be lost, thus making it impossible to fix incorrect under-segmentations over time.

We introduce here the notation $\Phi(l_j, o_m)$ to denote the pairwise count in the global map between a persistent segment label $l_j\in\mathcal{L}$ and a persistent instance label $o_m\in\mathcal{O}$.
$\Phi(l_j, o_m)$ is used here to determine the mapping $I_t(o_i) = o_m$ from instance labels $o_i\in \mathcal{O}_t$ to instance labels $o_m\in \mathcal{O}$.
Specifically, for each segment $s_i\in \mathcal{S}_t$ with a corresponding $o_i \neq 0$ and no $I_t(o_i)$ defined yet, the persistent object label $\hat o_m$ with the highest pairwise count $\Phi(L_t(s_i), o_j) > 0$ is identified.
The object label $o_i$ is then mapped to $\hat o_m$ as $I_t(o_i) = \hat o_m$.
Remaining $o_i$ with no mapping $I_t(o_i)$ found are assigned a new persistent instance label $o_{new}$ as $I_t(o_i) = o_{new}$.
Following a similar reasoning as above, multiple labels $o_i\in \mathcal{O}_t$ are prevented from mapping to the same persistent label $o_m\in \mathcal{O}$ in order not to discard valuable instance segmentation information from the current frame.

The result of this data association step is a set of 3D segments $s_i\in\mathcal{S}_t$ from the current frame, each assigned a persistent segment label $l_j = L(s_i)$. 
Further, the corresponding object instance label is matched to a persistent label $o_m = I_t(o_i)$.
Additionally, each segment $s_i\in\mathcal{S}_t$ is associated with the semantic object category $c_i$ predicted by Mask R-CNN (Section \ref{ssec:seg_refinement}).

\subsection{Map integration}
\label{ssec:integration}
The 3D segments discovered in the current frame, including some which are enriched with class and instance information, are fused into a global volumetric map.
To this end, the Voxblox~\cite{oleynikova2017voxblox} \acronym{tsdf}-based dense mapping framework is extended to additionally encode object segmentation information.
After projecting the segments into the global \acronym{tsdf} volume using the known camera pose, voxels corresponding to each projected 3D point are updated to store the incoming geometric segment label information, following the approach introduced in~\cite{Furrer2018IncrementalOD}.
Additionally, for each $s_i\in\mathcal{S}_t$ integrated into the map at frame $t$ with corresponding $o_i \neq 0$, the pairwise count between $l_j = L(s_i)$ and the object instance $o_m = I_t(o_i)$ and the pairwise count between $l_j$ and the class $c_i$ are incremented as,
\begin{align}
     \Phi(l_j, o_m) &=  \Phi(l_j, o_m) + 1\enspace\\
     \Psi(l_j, c_i) &= \Psi(l_j, c_i) + 1\enspace .
\end{align}
Each 3D segment $s_j\in\mathcal{S}$ in the global map volume is then defined by the set of voxels assigned to the persistent label $l_j$.
If the segment represents a recognized, semantically annotated instance then it is also associated with an object label $\hat o_m = \arg\max_{o_m}\Phi(l_j, o_m)$ and a corresponding semantic class $\hat c_j = \arg\max_{c_j}\Psi(l_j, c_j)$.


\begin{figure*}[ht]
\centering
\subfloat{\includegraphics[width=\textwidth]{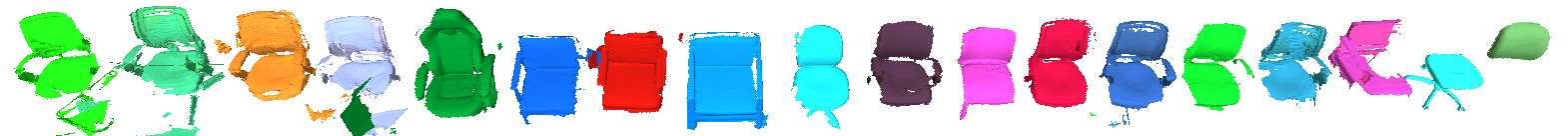}}\\[-2pt]
\subfloat{\includegraphics[width=\textwidth]{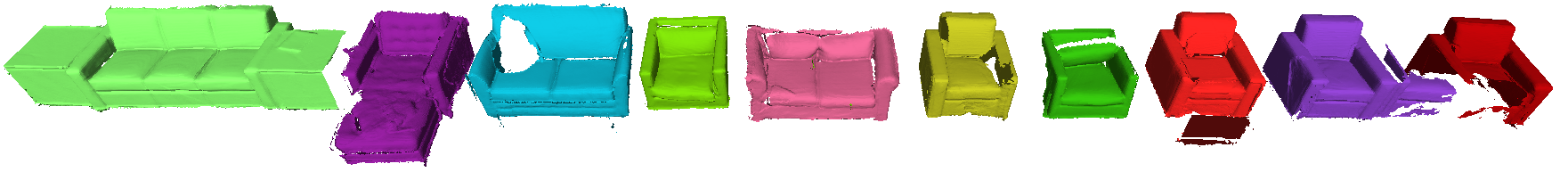}}\\[-2pt]
\subfloat{\includegraphics[width=\textwidth]{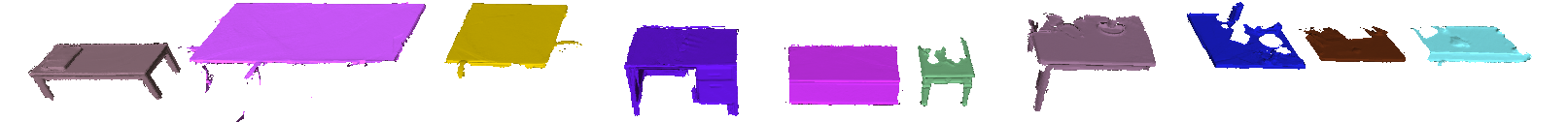}}\\[-7pt]
\subfloat{\includegraphics[width=\textwidth]{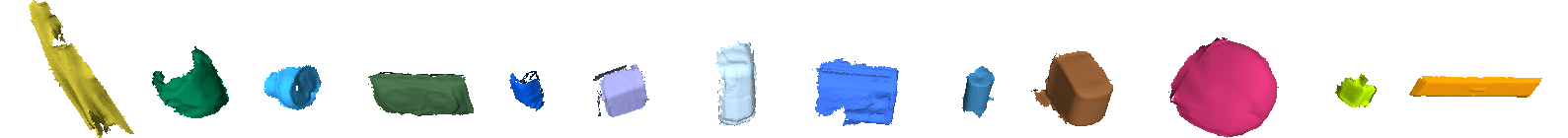}}\\
\caption{
Sample inventory of scene objects discovered during reconstruction of 10 indoor sequences from the SceneNN\mbox{\cite{7785081}} dataset. By virtue of a combined geometric-semantic segmentation scheme, the proposed mapping framework is able to detect recognized elements from a set of known categories and simultaneously discover novel, previously unseen objects in the scene. Accordingly, the shown collection features selected elements of predicted class \textit{chair} (first row), \textit{couch} (second row), and \textit{table} (third row), as well as a set of newly discovered objects without an associated semantic label (fourth row). Namely, the discovered objects correspond to (from left to right): a jacket, a plastic bag, two types of fans, a loudspeaker, a cardboard box, a computer case, a heater, a tissue paper roll, a kitchen appliance, a pillow, a tissue box, and a drawer.
The individual models, shown here in the form of meshes, densely describe the reconstructed object shapes and provide detailed smooth surface definitions.}
\label{fig:inventory}
\vspace{-4mm}
\end{figure*}

\section{EXPERIMENTS}
The proposed approach to incremental instance-aware semantic mapping is evaluated on a Lenovo laptop with an Intel Xeon E3-1505M eight-core CPU at 3.00\,GHz and an Nvidia Quadro M2200 GPU with 4\,GB of memory only used for the Mask R-CNN component.
The Mask R-CNN code is based on the publicly available implementation from Matterport,\footnote{\href{https://github.com/matterport/Mask_RCNN}{https://github.com/matterport/Mask\_RCNN}} with the pre-trained weights provided for the Microsoft COCO dataset~\cite{COCO}.
In all of the presented experimental setups, maps are built from RGB-D video with a resolution of 640x480 pixels.

To compare against previous work in~\cite{PhamInstances}, we evaluate the 3D segmentation accuracy of the proposed dense object-level semantic mapping framework on real-world indoor scans from the SceneNN~\cite{7785081} dataset, improving over the baseline for most of the evaluated scenes.
A sample inventory of object models discovered in these scenes is shown to contain recognized, semantically annotated elements, as well as newly discovered, previously unseen objects.
Lastly, we report on the runtime performance of the proposed system.

The framework is further evaluated within an online setting, mapping an office floor traversed by a robotic platform.
Although the system operates at only 1\,Hz, qualitative results in the form of a semantically annotated object-centric reconstruction validate the online nature of the approach and show its benefits in real-world, open set conditions.

\subsection{Instance-aware semantic segmentation}
Several recent works explore the task of semantic instance segmentation of 3D scenes.
The majority of these, however, take as input the full reconstructed scene, either processing it in chunks or directly as a whole.
Because such methods are not constrained to progressively fusing predictions from partial observations into a global map but can learn from the entire 3D layout of the scene, these are not directly comparable to the approach presented in this work.
Among the frameworks that instead explore online, incremental instance-aware semantic mapping, the work in~\cite{PhamInstances} is, to the best of our knowledge, the only one to present quantitative results in terms of the achieved 3D segmentation accuracy. 
While a comparison with~\cite{PhamInstances} does not provide any insight into the performance of the proposed unsupervised object discovery strategy, it can help to assess the efficacy of the semantic instance-aware segmentation component of our system.

In their work, Pham \textit{et al.}~\cite{PhamInstances} report instance-level 3D segmentation accuracy results for the NYUDv2 40 class task, which includes commonly-encountered indoor object classes, as well as structural, non-object categories, such as \textit{wall},\textit{ window}, \textit{door}, \textit{floor}, and \textit{ceiling}.
This set of classes is well-suited for semantic segmentation tasks in which the goal is to classify and label every single element, either voxel of surfel, of the 3D scene.
Indeed, the approach in~\cite{PhamInstances} initially employs a purely semantic segmentation strategy, and later clusters the semantically annotated scene into individual instances.
However, a set of classes which includes non-object categories does not apply to the  object-based segmentation approach proposed in this work.
Therefore, rather than training on a class-set that does not meet the requirements and goals of the proposed framework, we relied on a Mask R-CNN model trained on the 80 Microsoft COCO object classes~\cite{COCO}.
We then evaluated the segmentation accuracy on the 9 object categories in common between the NYUDv2 40 COCO class tasks.
Specifically, we picked the 9 categories that have an unambiguous one-to-one mapping between the two sets.

\begin{figure*}[!ht]
\centering
\setlength{\tabcolsep}{0.2\tabcolsep}
\subfloat[\label{fig:robot}]{ \includegraphics[width=0.41\columnwidth]{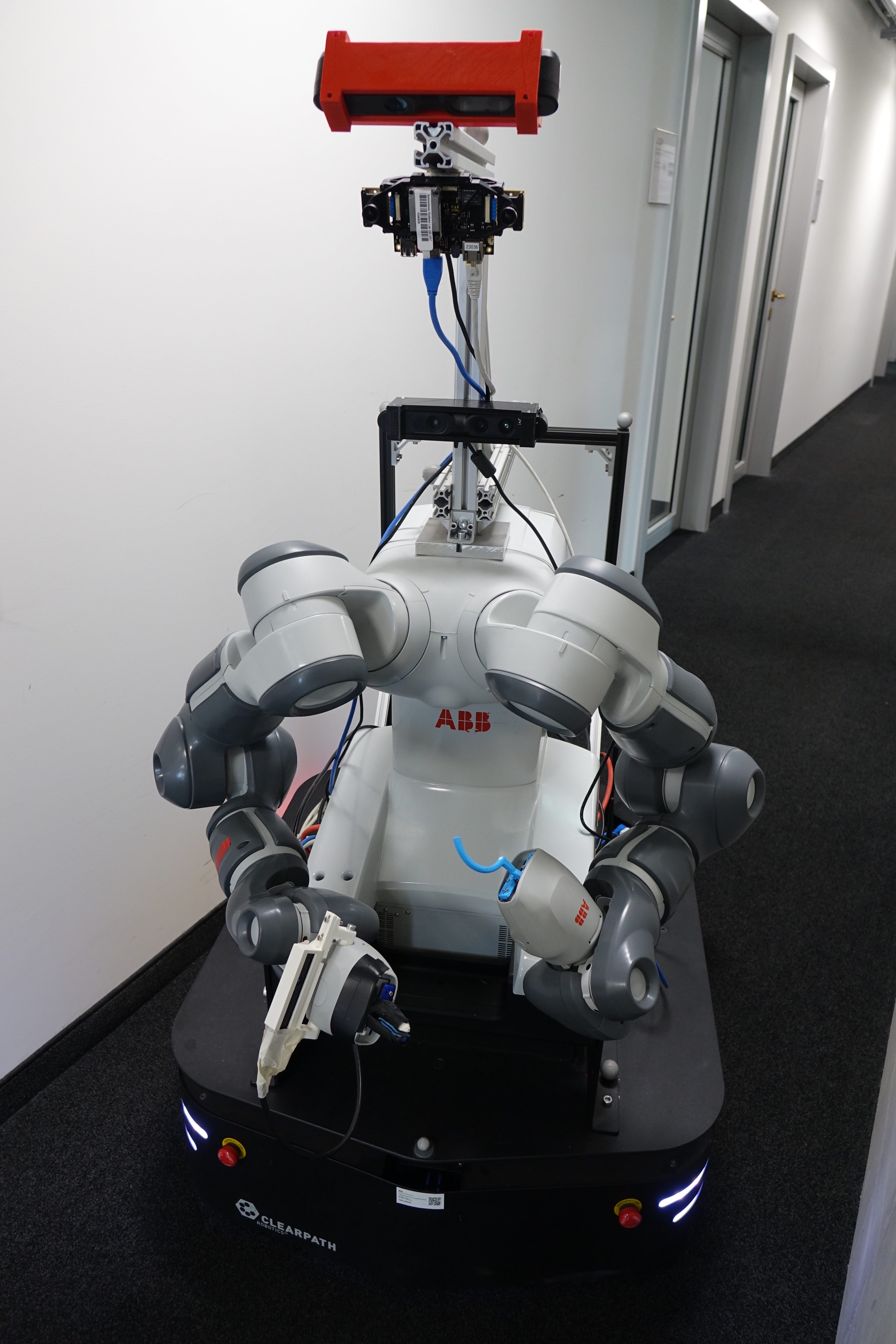}} \hspace{5.5mm}
\subfloat[\label{fig:mombifull}]{\includegraphics[width=1.39\columnwidth]{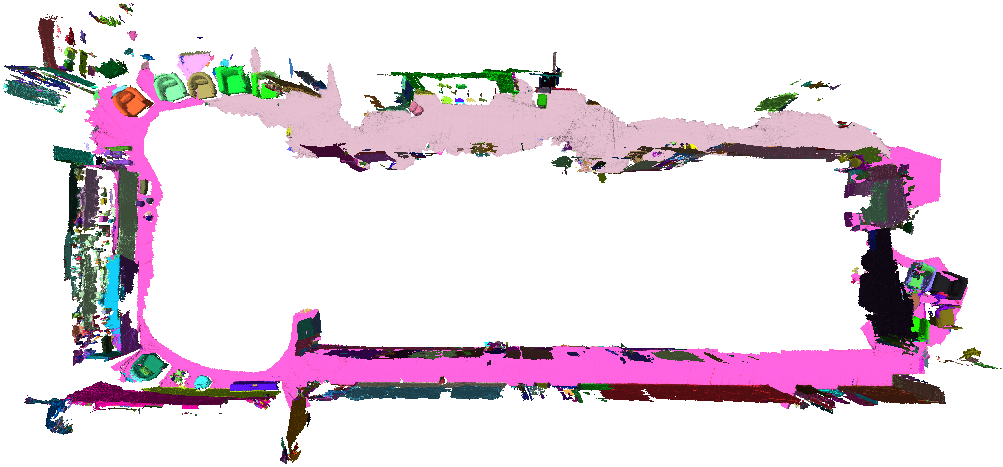}}\\
\subfloat[\label{fig:mobmimap}]{\includegraphics[width=0.65\columnwidth]{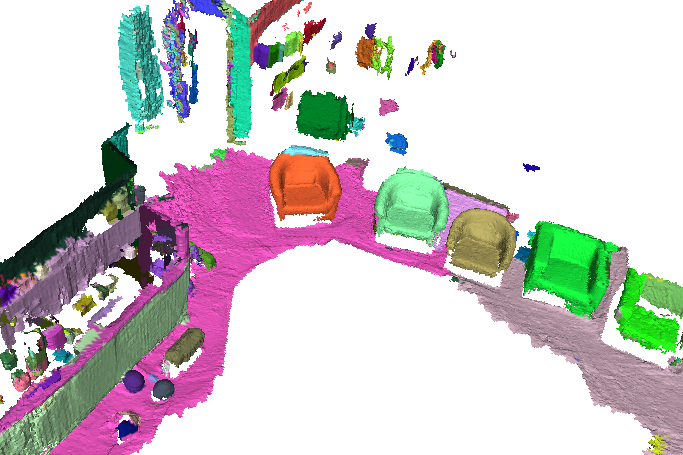}} \hspace{0.1mm} \subfloat[\label{fig:mobmisemantic}]{\includegraphics[width=0.65\columnwidth]{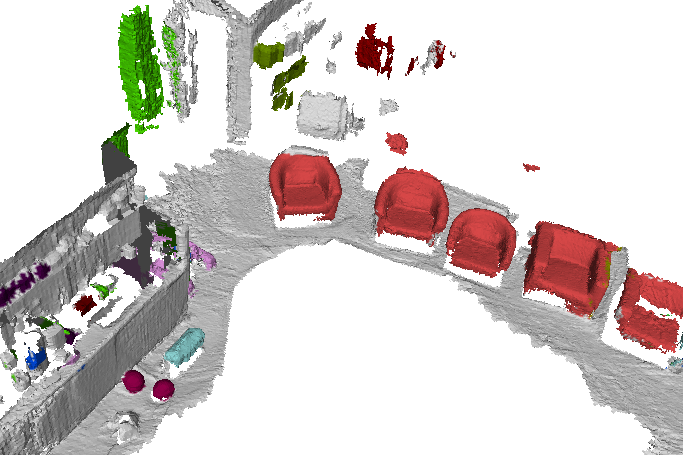}} \hspace{0.1mm}
\subfloat[\label{fig:mobmislices}]{\includegraphics[width=0.65\columnwidth]{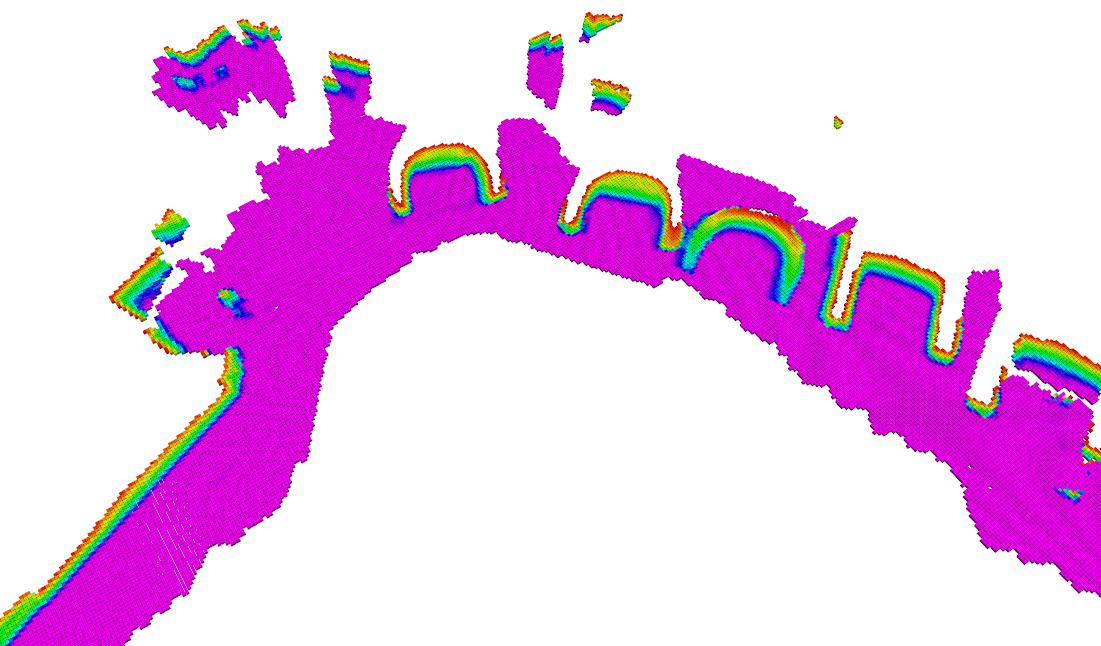}}
\caption{
Figure~\protect\subref{fig:robot} shows the robotic platform used for the online mapping experiment of an office floor.
The map is reconstructed from RGB-D data recorded with two Primesense cameras mounted on an ABB YuMi robot attached to a Clearpath Ridgeback mobile base.
The final map shown as a mesh in Figure~\protect\subref{fig:mombifull} is reconstructed at a voxel size of 2\,cm.
Figure~\protect\subref{fig:mobmimap} shows a detail of the map where individual objects identified in the scene are represented with different colors.
The corresponding semantic categories of the recognized instances are shown in Figure~\protect\subref{fig:mobmisemantic} using the same color coding as in Figure~\protect\ref{fig:teaser}.
Figure~\mbox{\protect\subref{fig:mobmislices}} shows a single horizontal slice at 1\,m height of the reconstructed \mbox{\protect\acronym{tsdf}} grid with magenta indicating observed free space, knowledge about which can directly benefit safe planning for navigation and interaction tasks.}
\label{fig:mobmi}
\vspace{-4mm}
\end{figure*}

The proposed approach is evaluated on the 10 indoor sequences from the SceneNN~\cite{7785081} dataset for which~\cite{PhamInstances} reports instance-level segmentation results.
For each scene, the per-class \acronym{ap} is computed using an \acronym{iou} threshold of 0.5 over the predicted 3D segmentation masks.
As~\cite{PhamInstances} only provides class-averaged \acronym{map} values, these are compared with \acronym{map} averaged over the 9 evaluated categories.
The results in Table \ref{tab:evaluation} show that the proposed approach outperforms the baseline on 7 of the 10 evaluated sequences, however it is worth noting again that the reported \acronym{map} values are computed over a smaller set of classes.

Besides evaluating the semantic instance-aware segmentation, Figure~\mbox{\ref{fig:inventory}} additionally shows a sample inventory of selected object instances detected and densely reconstructed across the 10 sequences.
Along with recognized, semantically annotated objects, the shown collection includes newly discovered scene elements, highlighting the benefits of the proposed unsupervised object discovery strategy.

Table~\ref{tab:runtime} shows the running times of the individual components of the framework averaged over the 10 evaluated sequences.
The numbers indicate that the system is capable of running at approximately 1\,Hz on 640x480 input.

\renewcommand{\arraystretch}{1.2}
\begin{table}[H]
\caption{
Measured execution times of each stage of the proposed incremental object-level mapping framework, averaged over the 10 evaluated sequences from the SceneNN~\cite{7785081} dataset with RGB-D input of 640x480 resolution.
Inference through Mask R-CNN runs on GPU, while the remaining stages are implemented on CPU. The map resolution is set here to 1\,cm voxels. Note that the components with * can be processed in parallel.}
\centering
\begin{tabular}{lc}
\toprule
\multicolumn{1}{l}{Component} & Time (ms)           \\ 
\midrule
Mask R-CNN *                    & 407  \\ 
Depth segmentation *            & 753       \\
Data association              & 136             \\
Map integration              & 276       \\ 
\bottomrule
\end{tabular}
\label{tab:runtime}
\vspace{-2mm}
\end{table}

\subsection{Online reconstruction and object mapping}
The proposed system is evaluated in a real-life online mapping scenario.
The robotic setup used for evaluation consists of a collaborative dual arm ABB YuMi robot mounted on an omnidirectional Clearpath Ridgeback mobile base.
The platform is equipped with the custom-built visual-inertial sensor described in\mbox{\cite{nikolic2014synchronized}}, used only for online localization.
Two PrimeSense RGB-D cameras are mounted facing forwards and downwards at 45 degrees, respectively, to capture dense depth maps and color images at an increased effective field of view.
%
%
%
%
The complete setup is shown in Figure~\ref{fig:robot}. 

Within the course of 5 minutes, the mobile base was manually steered along a trajectory through an entire office floor.
Real-time poses were estimated through a combination of visual-inertial and wheel odometry and online feature-based localization in an existing map built and optimized with Maplab~\cite{Schneider2018MaplabAO}.
During scanning, the RGB-D stream of the two depth cameras is recorded to be later fed through our mapping framework at a frame rate of 1\,Hz, emulating real-time on-board operation.
That is, any frames that exceed the processing abilities of the system are discarded and not used to reconstruct the object-level map of the scene.
The accompanying video illustrates the progressive output of the incremental reconstruction and segmentation of the scene.

Qualitative results for the final object-centric map are shown in Figure~\ref{fig:mobmi}.
%
%
Despite only a subset of the incoming RGB-D frames being integrated into the map volume, the resulting reconstruction of the environment densely describes the observed surface geometry.
The system is further able to detect recognized objects of known class, and to discover novel, previously unseen object-like elements in the scene.
Reconstructed over a trajectory length of over 80\,m with a voxel resolution of 2\,cm, the entire map fits into 605\,MB of memory, which is comparable with the memory usage of the bare Voxblox framework.
The final volumetric map additionally provides free space information, relevant for safe planning for robotic navigation and interaction tasks.
Such tasks can be carried out in parallel, as the total computational load of the individual components of the framework corresponds to using only 5 out of the 8 CPU cores.

It is worth noting that the quality of the reconstruction in Figure~\mbox{\ref{fig:mobmi}} has been in part affected by empirically measured pose estimation errors accumulating up to 0.5\,m.
Because this work focuses entirely on mapping and assumes localization to be given, we leave the task of quantifying the impact of inaccurate localization on the map quality to future work.


\section{CONCLUSIONS}
We presented a framework for online volumetric instance-aware semantic mapping from RGB-D data.
By reasoning jointly over geometric and semantic cues, a frame-wise segmentation approach is able to infer high-level category information about detected and recognized elements, and to discover novel objects in the scene, for which no previous knowledge about their exact appearance is available.
The partial segmentation information is incrementally fused into a global map and the resulting object-level semantically annotated volumetric maps are expected to directly benefit both navigation and manipulation planning tasks.

Real-world experiments validate the online nature of the proposed incremental framework.
However, to achieve real-time capabilities, the runtime performance of the individual components requires further optimization.
A future research direction involves investigating the optimal way to fuse RGB and depth information within a unified per-frame object detection, discovery and segmentation framework.






\section*{ACKNOWLEDGMENT}

The authors would like to thank T. Aebi for his help in collecting data for the office floor mapping experiment.


\bibliographystyle{IEEEtran}
\bibliography{IEEEabrv,IEEEexample}

\end{document}